%% file: main.tex
\pdfoutput=1

\documentclass[11pt]{article}

\usepackage{acl}

\usepackage{times}
\usepackage{latexsym}

\usepackage[T1]{fontenc}

\usepackage[utf8]{inputenc}
\usepackage{graphicx}
\usepackage{microtype}

\usepackage{amsmath}
\usepackage{amsfonts}
\usepackage[utf8]{inputenc}
\usepackage{xcolor}
\usepackage{colortbl}

\newcommand{\comment}[1]{}

\DeclareMathOperator*{\E}{\mathop{\mathbb{E}}}

\DeclareMathOperator*{\argmax}{argmax}

\title{Towards Computationally Verifiable Semantic Grounding for Language Models}

\author{Chris Alberti \;\;\; Kuzman Ganchev \;\;\; Michael Collins \\ {\bf Sebastian Gehrmann} \;\;\; {\bf Ciprian Chelba}\\
  Google Research\\
  \texttt{\normalsize \{chrisalberti,kuzman,mjcollins,gehrmann,ciprianchelba\}@google.com}
  }
  
\begin{document}
\maketitle

\begin{abstract}

The paper presents an approach to semantic
grounding of language models (LMs) that conceptualizes the LM as a conditional model
generating text given a desired semantic message formalized as a set of entity-relationship
triples. It embeds the LM in an auto-encoder
by feeding its output to a semantic parser
whose output is in the same representation
domain as the input message.
Compared to a baseline that generates text using greedy search, we demonstrate two techniques that improve the fluency and semantic accuracy of the generated text: The first technique samples multiple candidate text sequences from which the semantic parser chooses. The second trains the language model while keeping the semantic parser frozen to improve the semantic accuracy of the auto-encoder. 
We carry out experiments on the English WebNLG 3.0 data set, using BLEU to measure the fluency of generated text and standard parsing metrics to measure semantic accuracy. We show that our proposed approaches significantly improve on the greedy search baseline. Human evaluation corroborates the results of the automatic evaluation experiments.

\end{abstract}

\section{Introduction}

A statistical language model (LM) in its standard formulation assigns a probability to sequences of tokens that constitute an (ideally lossless) representation of text units such as sentences, paragraphs or larger. The original use for LMs was as prior probability in source-channel models for automatic speech recognition, machine translation and other similar tasks. Such use cases for LMs have largely disappeared\footnote{We exclude encoder-only masked "language models" such as BERT from our discussion, as their use to generate text is limited~\citep{DBLP:journals/corr/abs-1905-12790}.
} with the shift to seq2seq models \citep{sutskever2014sequence} as conditional LMs (direct models) for text given speech, or text in foreign language, etc.

Very large LMs trained on huge amounts of text have demonstrated bewildering capabilities in dealing with a large array of natural language processing and understanding tasks, e.g. GPT-3 \citep{brown2020language}, PaLM \citep{chowdhery2022palm}, Gopher \citep{rae2021scaling}, Lamda \citep{thoppilan2022lamda}. Displaying near flawless fluency, the text sampled from LMs turns out to also be semantically adequate to an ad-hoc prompt setup in one or a few ``shot'' manner on a wide range of NLP tasks \citep{brown2020language}. This is surprising since the LM is trained strictly as a generative model without any specific context other than the preceding text; as pointed out by many, the text sampled from LMs trained on large amounts of surface text cannot be expected to come with any weak or strong guarantees in terms of semantic adequacy, e.g.~\newcite[][Sec.~5]{Bender2021StochasticParrots}.

In this work, we approach semantic grounding of language models (LMs) by conceptualizing the LM as a conditional model generating text given a desired semantic message formalized as a set of entity-relationship triples. The LM is embedded in a semantic auto-encoder by feeding its output to a semantic parser whose output is in the same representation domain as the input message.

We propose two techniques for improving the semantic adequacy of the text generated by the LM. We find that both approaches improve the semantic $F1$ score significantly over the greedy baseline while preserving the same text fluency as measured using BLEU or METEOR scores against the reference text for a given $S$. 

In order to mitigate the limitations of the semantic parser (SP) in our evaluation, we also conduct human evaluations.  We collect a high recall set of possible meaning triples represented by a piece of text and ask human raters to evaluate whether each one is present in the text.  Starting with a high-recall set of triples mitigates recall failures of the SP, while collecting human ratings mitigates precision failures in the high-recall set.  Human evaluation corroborates the conclusions from experiments using automatic metrics, showing that our proposed techniques lead to a small but significant improvement in semantic grounding of generated text.

\section{Background and Intuition}

The use case for decoder-only auto-regressive LMs---input text (``prompt'') is fed to the LM and then output text is sampled from the LM given the state induced by the ``prompt''---follows the original use of seq2seq models~\citep{sutskever2014sequence} and is best described by conceptualizing the LM as a conditional model $P(W|\mathrm{prompt})$, a probability distribution on output text $W$ given an input textual prompt.

In our work we take a similar view on LMs, framing them as conditional models whose purpose is to encode meaning $S$ as text $W$, $P(W|S; \theta)$, except that the semantic message $S$ is formally defined and different from surface form text. Communication thus entails the exchange of semantic messages $S_1, \ldots S_n$ between conversation partners, or reader and writer, encoded as utterances $W_1, \ldots, W_n$. Besides possessing a LM that is used to verbalize the semantic message $S$ into words $W$, each speaker is able to ``decode'' a semantic meaning $S^+ = \argmax_{S} R(S|W; \phi)$ using a semantic parser (SP).


In this view, training a LM from surface text alone is no longer possible. To be able to do so we first need a SP that is able to recover the semantic message $S$ in a unit of text $W$. 
Assuming the SP is available, we can use it to generate training pairs $(S,W)$ for a semantic LM $P(W|S; \theta)$. The question whether such a LM is {\em semantically grounded} is now well posed: we can use held-out semantic messages $S^*$ and 
then sample text $W^*$ from our LM $P(W|S^*; \theta)$ and compare the output $S^+ = \argmax_{S} R(S|W^*; \phi)$ of the SP to the desired input message $S^*$ and thus check the semantic accuracy of text generated by the LM.

However, the exact definition of semantic messages $S$ and finding a SP that is able to extract them from unrestricted text is infeasible with current methods. For a proof of concept we settle on using the highly constrained setup in the WebNLG challenge~\citep{castro-ferreira-etal-2020}. In this setup, a large text-to-text seq2seq model such as T5~\citep{raffel2019T5} is incrementally trained as a WebNLG SP and its accuracy in producing entity-relationship (E-R) triples as defined in WebNLG is measured on available test data. 
Using this SP and a semantic LM $P(W|S; \theta)$, also bootstrapped from T5 and incrementally trained on the WebNLG training data, we can measure the extent to which the LM produces "semantically grounded" text by computing Precision and Recall between the input set of E-R triples fed to the LM and that output by running the SP on the text sampled from the LM. 

In the absence of a perfect SP (with Precision/Recall = 1.0) we can no longer computationally verify the semantic grounding of the LM: a SP with Recall < 1.0 misses E-R triples present in the text generated by the LM and we can no longer strictly guarantee its semantic adequacy. We do note however that a SP that does come close to Recall = 1.0 at a Precision < 1.0 value can still guarantee semantic adequacy of the text as long as the set of E-R triples produced is a subset or equal to the desired input set.

Our work evaluates the semantic adequacy of LM generated text 
$W^* = \textsc{generate}(P(W|S^*))$ and investigates algorithms that improve it while preserving fluency. 

A family of simple inference-time techniques consists of sampling different word sequences in ways that preserve fluency and picking the one that produces the highest semantic $F1$ score with respect to the input $S$, without re-estimation of the LM parameters. Another approach conceptualizes the LM followed by the SP as a semantic auto-encoder and estimates the LM while keeping the SP ``frozen'' such that the output sequence $W^*$ maximizes the probability of the correct semantic parse $R(S^*|W^*)$. The latter estimation approach can be applied in both training and at inference time; in this work we only investigate the latter.

\input{related_work}

\input{model}

\input{dataset}

\input{results}

\section{Conclusions}

We have presented an approach to semantic grounding of language models that conceptualizes the LM as a conditional model generating text given a desired semantic message and embeds it in a semantic auto-encoder model by feeding the LM output to a semantic parser whose output is in the same representation domain as the input message. 

We evaluate a simple baseline that generates text using greedy search and show that one can improve both the fluency of the generated text and its semantic accuracy by two simple techniques. The first is a sampling one that generates a few candidate text sequences and lets the semantic parser choose the better one. The second one trains the language model (while keeping the semantic parser frozen) with the aim of improving the auto-encoder semantic accuracy. 
We carry out experiments on the English WebNLG 3.0 data set, using BLEU to measure the fluency of generated text and standard semantic parsing metrics to measure the match between the output of the parser and the desired (input) semantic message. We show that our proposed approaches improve the semantic accuracy of generated text significantly over the greedy search baseline but they are not entirely additive. Human evaluation corroborates the results of automatic evaluation experiments.

\section{Limitations and Future Work}

The main challenge of this work was finding a suitable representation for the desired semantics of the generated text along with a semantic parser that could generate such representations from unconstrained text. The suitability of RDF triples is questionable for open domain semantics representation in unconstrained text, so an approach aiming at widening the scope of the current work would have to first address this obstacle.

Assuming the parser operating point on the Precision/Recall curve can be brought close enough to perfect Recall, the framework proposed would lend itself to computationally verifiable semantic grounding, namely being able to guarantee the fact that the generated text does not convey meaning outside the intended one.



\bibliography{anthology,custom}

\appendix

\section{Human Evaluation Template}
\label{sec:appendix_human_eval_template}

\begin{figure*}
    \centering
    \fbox{
    \includegraphics[width=15cm]{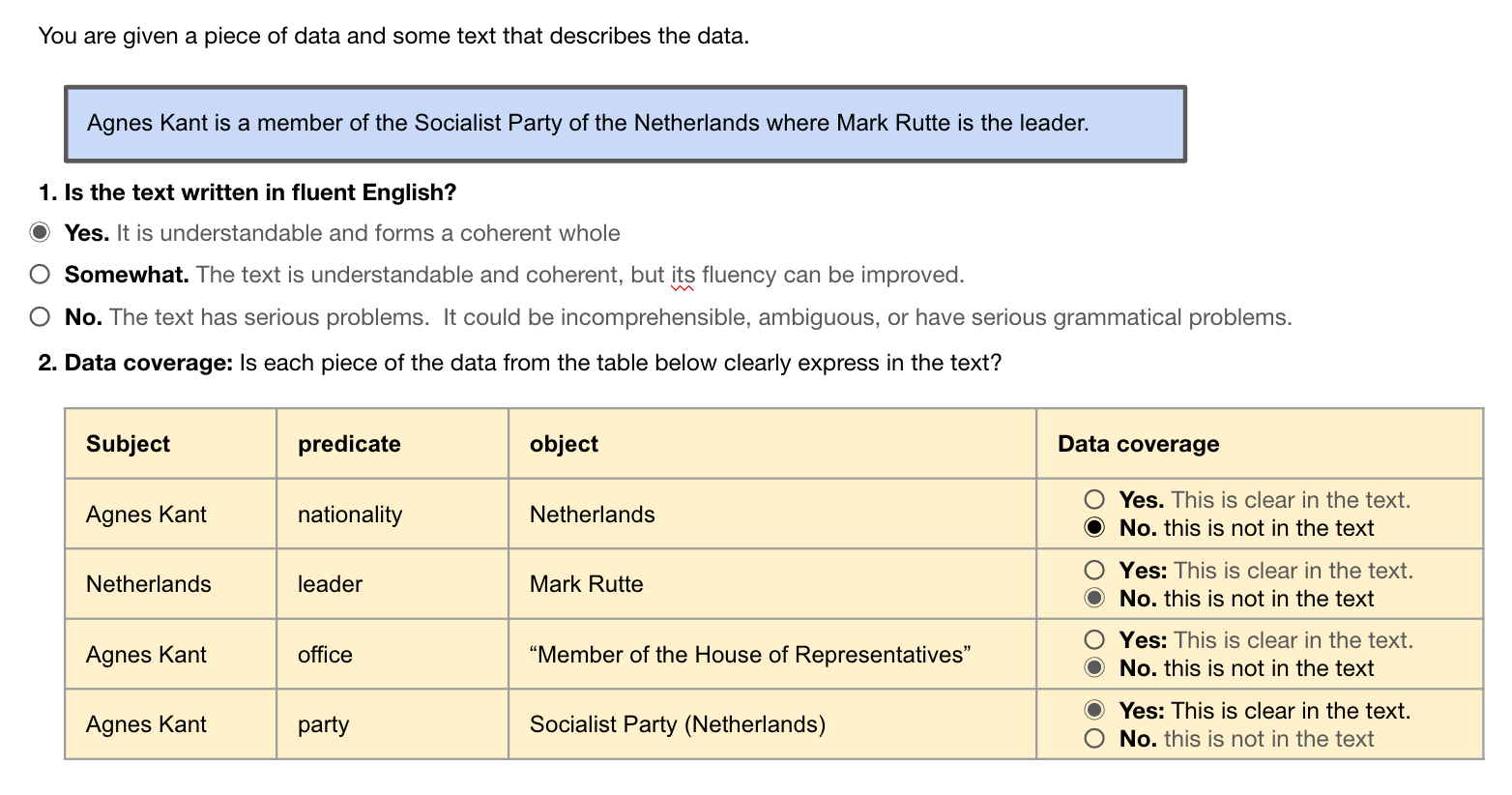}
    }
    \caption{The fluency and data coverage questions from our evaluation template.}
    \label{fig:screenshot_q1_q2}
\end{figure*}

\begin{figure*}
    \centering
    \fbox{
    \includegraphics[width=15cm]{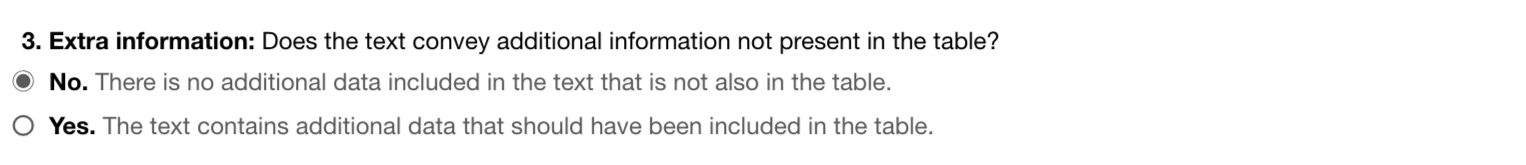}
    }
    \caption{A screenshot of the question asking annotators if the triples cover all the information in the text.}
    \label{fig:screenshot_q3}
\end{figure*}

A screenshot of the human evaluation template is shown in Figure~\ref{fig:screenshot_q1_q2} and Figure~\ref{fig:screenshot_q3}.

\end{document}

%% file: related_work.tex
\section{Related Work}

The idea of building auto-encoders for language where the latent variable itself is language was previously explored by \newcite{miao2016language}, where the authors propose an inference approach for sentence compression based on variational auto-encoders. Unlike  \newcite{miao2016language}, our work draws the latent representation from a probability distribution conditioned on the input semantic parse, rather than a background language model. The optimization strategy is also different since we perform a small number of gradient descent steps instead of employing variational inference.

\newcite{sun-etal-2018-logician} tackle a very similar problem in a reinforcement learning setup, training both LM ("narrator") and open information extraction models jointly. Our work fixes the semantic parser while still taking advantage of the ability to back-propagate the error of reconstructing the correct semantic message.

The approach of data-to-text-to-data is also a somewhat common evaluation strategy. Starting with \citet{wiseman2017challenges}, people have been using information extraction to match information in text with that of the input. More recent work like \citet{manning2021referenceless} tries to parse text into AMRs. In this work we go a step further and attempt to directly improve generation quality by sampling or by backpropagating parsing signals into the generation component.

%% file: model.tex
\section{Model}


Let $W$ be a text unit (a sentence in the WebNLG data) and $S$ be a representation of the semantic message of $W$ (a sequence of WebNLG triples). We define a semantic LM and a semantic parser as two sequence-to-sequence models:
\begin{itemize}
    \item $P(W|S;\theta)$: a \textit{semantic LM} trained to encode a semantic message $S$ into text units $W$;
    \item $R(S|W;\phi)$: a \textit{semantic parser (SP)}, trained to decode the semantic message $S$ from a text unit $W$.
\end{itemize}
Our objective is to develop a method that generates computationally verifiable text: given a semantic message $S^*$, we would like to find a verbalization $W^*$ such that running the semantic parser $R$ on $W^*$ will result in a semantic message $S^+$ as close as possible to the original $S^*$. In this work we compare a baseline to two methods designed to achieve this objective:  a sampling approach and a new approach we refer to as ``greedy finetuning''.

In the baseline approach we decode greedily with $P(W|S;\theta)$ to encode the input semantic message $S^*$ into the text unit $W^*$. In the sampling approach, we sample text units from the $P(W|S=S^*; \theta)$ model using temperature or nucleus sampling \cite{holtzman2019curious}, then run the semantic parser on every sampled text unit, finally select as $W^*$ the text unit that maximizes the $F1$ score between the semantic parse $S^+ = \argmax_{s} R(S|W^*;\phi)$ and the input semantic message $S^*$. 

In ``greedy finetuning'' we take a more complex approach and re-estimate the LM based on feedback from the semantic parser. We iteratively perform greedy decoding  with the semantic LM $P$ to obtain a text unit $W_i$, then adjust the semantic LM $P$ while keeping the SP model $R$ ``frozen'' in order to increase the probability of reconstruction of the correct semantic message. After $k$ steps, we pick the final text unit $W^*$ from the set $\{W_1, \ldots, W_k\}$, choosing the one that leads to the best F1 score between the decoded semantic message $S^+$ and the original message $S^*$.

The semantic LM update in ``greedy finetuning'' is performed by taking a gradient ascent step on the following objective function:
\begin{align*}
  J(S^*, S^+; \theta, \phi) = \E_{W \sim P(\cdot|S^*;\theta)} [R(S^+|W;\phi)],
  \label{eq:obj}
\end{align*}
where keeping the semantic parser frozen corresponds to keeping $\phi$ constant and only updating $\theta$.\footnote{We note that greedy finetuning is expensive since it requires updating all the parameters of the semantic LM $k$ times for every inference batch. The approach could be made substantially more efficient, however, by employing parameter efficient methods such as prompt tuning \cite{lester2021power}.}

Since both $P$ and $R$ are neural networks, the gradient ascent step for $\theta$ can \textit{mostly} be computed by backpropagation. However, the objective $J$ is not differentiable because it requires auto-regressively decoding from the model $P$ to sample values for $W$ and estimate the expectation of correct reconstruction. We get around this difficulty by sampling the text unit $W$ using greedy decoding and then employing the straight-through gradient estimator (\newcite{bengio2013estimating}, \newcite{hinton2012neural}). The estimator is illustrated in more detail in section \ref{sec:st-approx}.

In addition to (1) greedy (baseline), (2) sampling and (3) greedy finetuning, we consider two ensemble methods: (4) greedy+sampling and (5) greedy finetuning+sampling. Greedy+sampling will run both methods (1) and (2), and then pick the verbalization $W^*$ that leads to the reconstruction with the highest F1 score. Greedy finetuning+sampling will similarly pick the best output after running methods (2) and (3).


\subsection{Straight Through Approximation}
\label{sec:st-approx}
To propagate the gradient of $J$ through the SP into the language model we need to compute
\begin{equation*}
    \frac{\partial R(S|W^*;\phi)}{\partial \theta},
\end{equation*} 
where $W^* = \argmax P(W|S^+; \theta)$ is computed using greedy search $\textsc{g\_search}$.

Following \newcite{hinton2012neural}, during the forward pass we transmit the text tokens $W^*$, and during the backward pass we pretend as if for every token the real value distribution over the vocabulary was transmitted instead. For this purpose, we can regard the $W^*$ sequence passed to the SP as a binary tensor  
\begin{equation*}
W^* = \textsc{one\_hot}(\textsc{g\_search}(\log P(W|S^+; \theta)))
\end{equation*} 
of shape $(b, l, v)$ where $b$ is the batch size, $l$ is the maximum length in tokens or word-pieces for text sequences including padding, and $v$ is the size of the vocabulary. 
	
In the straight-through (ST) approximation with automatic differentiation, we replace the $W^*$ tensor in the computational graph with the following:
\begin{eqnarray*}
W^*_\mathrm{ST} &= & P(W^*|S^+; \theta) +\\
& & \mathrm{S\_G}[W^* - P(W^*|S^+; \theta)]
\end{eqnarray*}
where the $\mathrm{S\_G}[\cdot]$ stop-gradient operator is the familiar pass-through in the forward direction and none-shall-pass in back-propagation. Thus in the forward direction the computation will proceed with $W^*_\mathrm{ST} = W^*$, however in the backward direction the computation will behave as if $W^*_\mathrm{ST} = P(W^*|S^+; \theta)$.


  


%% file: dataset.tex
\section{Data Set}

The WebNLG corpus~\cite{gardent-etal-2017-creating} comprises of sets of triplets describing facts (entities and relations between them) and the corresponding facts rendered in form of natural language text. The corpus contains sets of up to 7 triplets alongside one or more reference texts for each set. As explained in Section 2 of~\citep{castro-ferreira-etal-2020}, the English dev set consists of categories and entities seen in the training data; the English test set (D2T) consists of a mixture of seen entities and categories, as well as unseen entities and 5 unseen categories (28\%, 22\% and 50\%, respectively), making it hard to expect correct text generation and/or semantic parsing from our models. Nevertheless, we experiment in this condition as well.

The dataset can be used for work in both natural language generation and the reverse task of triplets extraction. Its main use was for the WebNLG natural language generation challenge with the goal to map the sets of triplets to text, including referring expression generation, aggregation, lexicalization, surface realization, and sentence segmentation. 

The initial (2017) WebNLG shared task required participating systems to generate English text from a set of DBpedia triples~\cite{gardent-etal-2017-webnlg} whereas the more recent (2020) WebNLG 3.0 data set used for the WebNLG+ challenge~\citep{castro-ferreira-etal-2020} additionally includes generation into Russian and semantic parsing of English and Russian texts, encompassing four tasks: RDF-to-English, RDF-to-Russian, English-to-RDF and Russian-to-RDF.

We only use WebNLG 3.0 English data, starting from a pre-trained language model~\cite{raffel2019T5} and fine-tuning it on the WebNLG training data for either LM or semantic (RDF) parsing. Since the model uses a encoder-decoder architecture we feed the conditioning information (RDF triples when incrementally training the LM or natural language text when training the SP) to the encoder and then let the decoder generate the output sequence (natural language text or RDF triples, respectively). We note that the exact order of the RDF triples for a given reference text is a degree of freedom that we could experiment with.

%% file: results.tex
\definecolor{Gray}{gray}{0.9}

\section{Experiments}

\begin{table*}[t]
    \centering
    \begin{tabular}{l|ccccc} 
       Method                  & BLEU & METEOR & chrF++ & Triple P / R / F1 & Improved\\
       \hline
       greedy (baseline)                  & 0.64 &   0.46 &  0.76 & 0.93 / 0.87 / 0.90 & - \\
       \rowcolor{Gray}
       sampling                & 0.51 &   0.43 &  0.71 & 0.97 / 0.92 / 0.95 & - \\
       sampling (t=0.3, p=0.95) & 0.63 & 0.46  &  0.76 & 0.96 / 0.91 / 0.94 & - \\
       \rowcolor{Gray} greedy+sampling       & 0.60 &   0.46 &  0.75 & 0.98 / 0.93 / 0.95 & 0.20 \\
       greedy+sampling (t=0.3, p=0.95) & 0.63 & 0.47  &  0.76 & 0.96 / 0.91 / 0.94 & 0.14 \\
       \rowcolor{Gray}
       greedy finetuned & 0.62 &   0.46 &  0.76 & 0.96 / 0.91 / 0.93 & 0.13\\
       greedy f.+sampl. (t=0.3, p=0.95) & 0.63 & 0.46 & 0.76 & \textbf{0.97 / 0.92 / 0.95} & \textbf{0.17} \\
    \end{tabular}
    \caption{Automatic evaluation of our method on ``seen'' test data (dev split of English WebNLG 3.0). Grayed rows have BLEU scores more than 1\% below the baseline. The ``Improved'' column displays the ratio of generations with a higher triple F1 compared to the greedy baseline.}
    \label{tab:automatic_eval_dev}
\end{table*}

\begin{table*}[t]
    \centering
    \begin{tabular}{l|ccccc} 
       Method                  & BLEU & METEOR & chrF++ & Triple P / R / F1 & Improved \\
       \hline
       Amazon AI (2020 1st)    & 0.54  & 0.42 &  0.69 &  \\
       bT5                     &  0.52 & 0.41 & 0.68 &  \\
       \hline
       greedy (baseline)                 & 0.54 & 0.41  &  0.69 & 0.58 / 0.50 / 0.54 & - \\
       \rowcolor{Gray}
       sampling                & 0.44 & 0.39  &  0.64 & 0.66 / 0.58 / 0.62 & - \\
       sampling (t=0.3, p=0.95) & 0.53 & 0.41  &  0.68 & 0.65 / 0.57 / 0.60 & - \\
       \rowcolor{Gray} greedy+sampling         & 0.51 & 0.41  &  0.67 & 0.67 / 0.59 / 0.63 & 0.31 \\
       greedy+sampling (t=0.3, p=0.95) & 0.54 & 0.41  &  0.69 & 0.65 / 0.57 / 0.61 & 0.23 \\
       greedy finetuned & 0.53 & 0.41  &  0.68 & 0.65 / 0.57 / 0.61 & 0.20 \\
       greedy f.+sampl. (t=0.3, p=0.95) & 0.53 & 0.41 & 0.68 & \textbf{0.69 / 0.61 / 0.65} & \textbf{0.34} \\
    \end{tabular}
    \caption{Automatic evaluation of our method on ``mixed: seen and unseen'' test data (test split of English WebNLG 3.0). Grayed rows have BLEU scores more than 1\% below the baseline. The ``Improved'' column displays the ratio of generations with a higher triple F1 compared to the greedy baseline.}
    \label{tab:automatic_eval}
\end{table*}

\begin{table}
    \centering
    \resizebox{\columnwidth}{!}{
    \begin{tabular}{l|ccc} 
       Model & F1 & P & R \\
       \hline
        Amazon AI (Shanghai) & 0.67 & 0.69 & 0.69 \\
        bt5 & 0.68 & 0.67 & 0.70 \\
        Our T5 & 0.63 & 0.63 & 0.64
    \end{tabular}}
    \caption{Exact match parsing performance of our T5 semantic parser (SP) on Text2RDF English WebNLG.
    }
    \label{table:parsing_acc}
\end{table}



As semantic LM and semantic parser we train T5 XXL models on the training set split of WebNLG for 100 steps with a batch size of 256 examples. We then compare three main methods: (1) the greedy decoding baseline, (2) sampling with different temperatures and nucleus probability mass, (3) greedy finetuning. We additionally report ensembled of these methods. (1)+(2) is ``greedy+sampling'', where we perform both greedy decoding and sampling and then pick the generated text with highest semantic parsing F1 score. Similarly (2)+(3) is ``greedy finetuning+sampling''.

For the sampling approach we alway pick 4 samples, we set the temperature to either 1.0 or 0.3, and the nucleus probability mass to either 1.0 or 0.95. For the greedy finetuning approach we perform gradient descent steps  on $\theta$ with AdaFactor, with a learning rate of $10^{-3}$. We perform 4 iterations  and return the hypothesis with the best semantic reconstruction. Greedy finetuning is performed on batches of 64 examples.


\subsection{Automatic Evaluation}

As an automatic metric of semantic grounding, we measure the standard Precision, Recall and F1 metrics used for evaluating semantic parsing accuracy, fixing the input WebNLG triples as reference and the triples reconstructed by the semantic parser as hypothesis. As for the fluency of the text generated by the LM, we measure it using BLEU \cite{papineni2002bleu}, METEOR \citep{banerjee2005meteor} and chrF++ \cite{popovic2017chrf++} against the human references in the D2T data sets. We note that there are between one and five references for each RDF triple, frequently two or three. 

The results of automatic evaluation on the English section of WebNLG 3.0 are shown in Tables \ref{tab:automatic_eval_dev} and \ref{tab:automatic_eval}. We first note that our greedy baseline matches the performance of the 2020 WebNLG winning system on BLEU and chrF++. Sampling and greedy finetuning increase triple F1 by 6\% and 7\% absolute on the test set, while keeping the BLEU score and other metrics within one point from the greedy baseline, showing that the semantic match with the desired input meaning can be significantly increased at minimal cost in the fluency of the generated text. The greedy finetuned+sampling ensemble gives us the highest performance, increasing triple F1 by 11\%, again keeping the fluency metrics within 1\% from the greedy baseline. The same overall trends can be observed for the WebNLG dev set in Table \ref{tab:automatic_eval_dev}. We additionally note that the sampling approach only performs well in this evaluation if temperature and nucleus probability mass are tuned.

We additionally report for both dev and test set the fraction of generations that with improved triple F1 compared to the greedy baseline. We only report the improved ratio for methods that are ensembled with greedy and so are guaranteed to always improve on automatically measured triple F1 compared to the baseline. We find that our best method, greedy finetuned+sampling, improves 34\% of our generations according to our automated metric of semantic verification.

As a sanity check for our semantic parser model, we report in Table \ref{table:parsing_acc} the performance of our T5 semantic parser on the official evaluation metrics of WebNLG Text2RDF. Since we only fine-tune T5 on the training split of WebNLG and we don't employ any additional techniques, the performance of our parser is competitive but lower than SotA systems.


\subsection{Human Evaluation}

\begin{table*}
    \centering
    \begin{tabular}{l|rrrrrr}
        Method             &  Examples &  Triples &  Fluency &  Precision &  Recall &  F-Score \\
        \hline
        reference         &     206   &   1278   &     1.00 &       0.77 &    0.91 &     0.84 \\
        \hline
        greedy (baseline) &     206   &   1278   & \bf 1.00 &       0.74 &    0.84 &     0.78 \\
        greedy + sampling &     115   &    719   &     0.99 &       0.74 &\bf 0.87 & \bf 0.80 \\
        greedy finetuned  &     112   &    675   &     0.98 &   \bf 0.76 &    0.84 & \bf 0.80 \\
    \end{tabular}
    \caption{Human evaluation results.  The row labeled ``reference'' has the evaluation of the human-generated reference text (provided as part of the test set).  Triples is the total number of triples rated, not the number of triples generated by the parser for a particular system. See text for details.}
    \label{tab:human_eval}
\end{table*}

Because our inference method relies on the SP, one could reasonably question whether it is appropriate to use the same SP also in the automatic evaluation. For example, it is possible that our method is learning to cheat by finding text with wrong semantics that the text-to-parse model incorrectly parses as the desired semantics.  In order to alleviate such concerns we augment the automatic evaluations with human ratings.

Ideally, expert annotators would encode the DBPedia semantics of reference text as well as system outputs. However, due to the size of DBPedia with tens of thousands of possible relations, it is infeasible for a human to perform the parsing task.  Instead, we present annotators with a text as well as candidate triples and ask them to evaluate whether each candidate triple is represented in the given text.  

We evaluate the text generated by our models in the following way: 
\begin{enumerate}
    \item Use the SP model to generate possible triples for the reference text as well as all model outputs. \label{step1}
    \item For the reference text and each system output, we present the raters with that text and the union of parse triples generated in step~\ref{step1}.
    \item For each triple, the raters verify whether it is implied by the text.
    \item We compute precision and recall for the reference text and each system with respect to the union of generated triples and the human ratings.
\end{enumerate}
We evaluate samples where the greedy decoding baseline did not produce the same result as greedy+sampling and greedy finetuned.  We see from Table~\ref{tab:automatic_eval_dev} that this happens in 23\% and 20\% of examples respectively.  We include the exact phrasing of the question as well as a screenshot of the rating interface in Appendix~\ref{sec:appendix_human_eval_template}.  In addition to the semantic annotation we also ask the raters to judge the fluency of the generated text. 

The results are shown in Table~\ref{tab:human_eval}.  The row labeled reference refers to the human-generated reference text which has a surprisingly low precision and recall.  The remaining rows have the same definitions as in Table~\ref{tab:automatic_eval_dev}.
We got 3-way annotations for 112 examples for greedy finetuned and 115 examples for greedy+sampling (and the corresponding 206 examples for greedy decoding).  Overall, we see that sampling improves recall while greedy finetuned improves precision, while not significantly changing the fluency of the text. Using bootstrap resampling on the examples, we find that greedy+sampling has significantly higher recall than greedy (p value 0.97), but greedy finetuned isn't better at the 95\% level in terms of precision (p value 0.948 with 10k samples).  The other metrics aren't significantly different.

In addition to corroborating the results from automatic evaluation, the human eval show strikingly low precision and recall of the ``reference'' text.  We show two illustrative examples of losses.  First, we display an example below of a recall loss, where the annotator did not adhere strictly to the semantics.

\vspace{1em}
\hspace{-1em}\begin{tabular}{@{}|p{0.94\columnwidth}|@{}}
    \hline
    \textbf{Input:} \small\texttt{Zaoyang | isPartOf | Hubei}  \\
    \small\texttt{Nie\_Haisheng | birthPlace | Zaoyang}  \\
    \hline
    \textbf{Target:} Nie Haisheng is from Zaoyang in Hubei province.\\
    \hline
\end{tabular}
\vspace{1em}

In this example, the rater chose to realize the \texttt{birthPlace} relation using the phrase \emph{is from}, but those were judged by our annotators as not having the same meaning. 

Below is an example of a precision error: 

\vspace{1em}
\hspace{-1em}\begin{tabular}{@{}|p{0.94\columnwidth}|@{}}
    \hline
    \textbf{Input:} \small\texttt{Zaoyang | isPartOf | Hubei} \\
    \small\texttt{Nie\_Haisheng | mission | Shenzhou\_6} \\
    \small\texttt{Nie\_Haisheng | birthPlace | Zaoyang} \\
    \hline
    \textbf{Target:} \\Born in Zaoyang city, Hubei, Nie Haisheng participated in the Shenzhou 6 mission.\\
    \hline
    \textbf{Implied:} \\
    \small\texttt{Nie\_Haisheng | birthPlace | Hubei} \\
    \hline
\end{tabular}
\vspace{1em}

In the example, {\footnotesize\texttt{Nie\_Haisheng | birthPlace | Zaoyang}} and {\small\texttt{Zaoyang | isPartOf | Hubei}} together imply {\small\texttt{Nie\_Haisheng | birthPlace | Hubeia}} and the SP picked up on this by generating the implied relation.  This was accepted by our annotators, and our metrics show it as a precision loss -- the generated text includes a triple that was not in the desired parse.\footnote{It may be possible in theory to enumerate implied relations by writing rules to avoid penalizing the text presented to the raters for including the semantics implied by the desired triples. However, this would make results more difficult to reproduce, and instead note that this is a shortcoming of the evaluation of the generated semantics.}